\documentclass[10pt,twocolumn,letterpaper]{article}

\usepackage[final]{cvpr}      % To produce the REVIEW version
\usepackage{xcolor}
\definecolor{cvprblue}{rgb}{0.21,0.49,0.74}

\usepackage[pagebackref,breaklinks,colorlinks,allcolors=cvprblue]{hyperref}
\usepackage{booktabs}

\def\paperID{} % *** Enter the Paper ID here
\def\confName{CVPR}
\def\confYear{2025}

\title{Visual-Linguistic Agent: Towards Collaborative Contextual Object Reasoning}
\author{Jingru Yang\\
Carnegie Mellon University\\
{\tt\small jingruy@andrew.cmu.edu}
\and
Huan Yu\\
Zhejiang University\\
{\tt\small 22225179@zju.edu.cn}
\and
Jingxin Yang\\
Beijing Institute of Petrochemical Technology\\
{\tt\small yangjx511@outlook.com}
\and
Chentianye Xu\\
Carnegie Mellon University\\
{\tt\small chentianye.xu@gmail.com}
\and
Biao Yin\\
TikTok\\
{\tt\small yinbiao15@outlook.com}
\and
Yu Sun\\
Sealand Technology Inc.\\
{\tt\small 18521326265@163.com}
\and
Shengfeng He\\
Singapore Management University\\
{\tt\small shengfenghe7@gmail.com}
}

\begin{document}

\teaser{
\centering
\small 
\begin{minipage}[t]{1\linewidth}
\centering
\includegraphics[width=1\columnwidth]{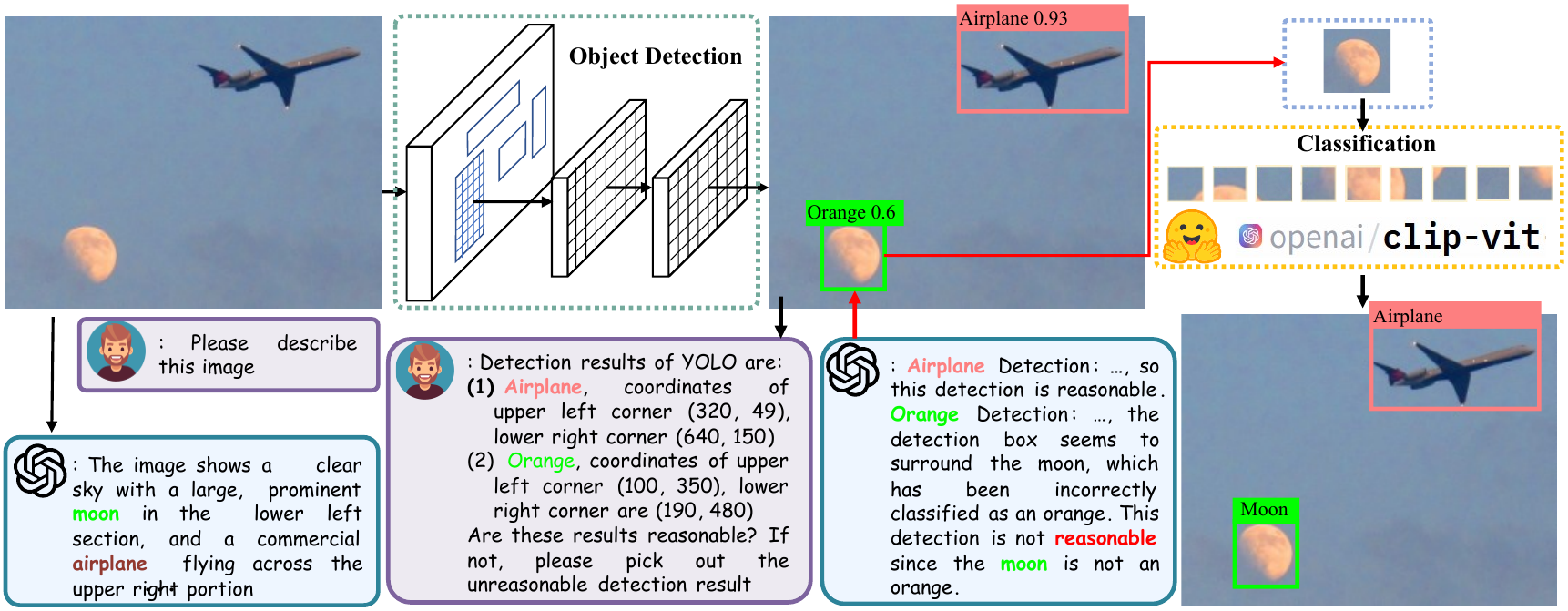}
\end{minipage}
\centering
\caption{An example from the proposed \textbf{Visual-Linguistic-Agent (VLA)} paradigm. Here, the Visual Agent (YOLO) detects objects in an image but misclassifies the moon as an orange. The Linguistic Agent (MLLM) evaluates these detection results and, using its reasoning capabilities, identifies and corrects the error. This example demonstrates the effectiveness of integrating visual detection with linguistic reasoning to enhance the accuracy and contextual reliability of object detection outcomes.
}
% \vspace{-10pt}
\label{FIG:att} 
}

\maketitle
\begin{abstract}
Multimodal Large Language Models (MLLMs) excel at descriptive tasks within images but often struggle with precise object localization, a critical element for reliable visual interpretation. In contrast, traditional object detection models provide high localization accuracy but frequently generate detections lacking contextual coherence due to limited modeling of inter-object relationships. To address this fundamental limitation, we introduce the \textbf{Visual-Linguistic Agent (VLA)}, a collaborative framework that combines the relational reasoning strengths of MLLMs with the precise localization capabilities of traditional object detectors. In the VLA paradigm, the MLLM serves as a central Linguistic Agent, working collaboratively with specialized Vision Agents for object detection and classification. The Linguistic Agent evaluates and refines detections by reasoning over spatial and contextual relationships among objects, while the classification Vision Agent offers corrective feedback to improve classification accuracy. This collaborative approach enables VLA to significantly enhance both spatial reasoning and object localization, addressing key challenges in multimodal understanding. Extensive evaluations on the COCO dataset demonstrate substantial performance improvements across multiple detection models, highlighting VLA’s potential to set a new benchmark in accurate and contextually coherent object detection.
\end{abstract}
    
\section{Introduction}
\label{sec:intro}

Object detection models have achieved remarkable success, particularly on large-scale datasets like COCO \cite{lin2014microsoft}. However, most existing object detection networks \cite{carion2020end, zhao2024ms, hu2024dac, zhang2022dino} primarily focus on generating high-quality proposals, employing separate branches to predict the class and bounding box for each proposal. While these models are effective in localizing objects with precision, they often overlook spatial and logical relationships between objects, resulting in detections that may lack contextual coherence, as shown in Figure \ref{FIG:att}. This gap highlights the need to incorporate spatial and logical relationship modeling to enhance the reasoning and overall accuracy of detection results. Despite its importance, this aspect has received relatively limited attention in the field.

Several studies \cite{liu2023cigar, zhao2023rgrn, xu2019spatial} have attempted to address this limitation by modeling spatial and logical relationships between objects using Graph Convolutional Networks (GCNs) \cite{liu2024rt} or Gated Recurrent Units (GRUs) \cite{chen2017spatial, zulqarnain2024efficient} applied to the visual representations of detected bounding boxes. These approaches primarily improve the bounding box regression branch by explicitly modeling relationships among objects. However, they are often constrained to relationships between local objects, neglecting the broader contextual environment and overall spatial structure among objects. Furthermore, as these methods typically operate independently of the classification branch, they provide limited improvements in the coherence and reasoning of final detection outcomes.

In contrast, multimodal large language models (MLLMs) \cite{li2024seed, liu2023visual_llava, li2023blip2, touvron2023llama, zhu2023minigpt4, driess2023palme} have demonstrated exceptional capabilities in modeling relationships across modalities, such as in image captioning \cite{nguyen2024improving} and visual question answering \cite{antol2015vqa}. MLLMs are proficient at constructing both spatial and logical relationships between objects, offering a more holistic understanding of visual scenes. However, despite their strong reasoning abilities, MLLMs struggle with precise object localization. Our motivation arises from the observation that combining the reasoning strengths of MLLMs with the robust localization capabilities of traditional object detection models can enhance both the accuracy and contextual coherence of object detection results.

In this paper, we propose the \textbf{Visual-Linguistic-Agent} (VLA), a collaborative framework that leverages the reasoning and relationship-modeling capabilities of MLLMs to enhance the contextual accuracy of traditional object detection models. In the VLA paradigm, the MLLM, which acts as a Linguistic Agent, collaborates with both an object detection \textbf{Vision Agent} and a classification \textbf{Vision Agent}. As illustrated in Figure \ref{FIG:att}, upon receiving a user request, the MLLM generates an image caption and evaluates detection outputs provided as text prompts from the Vision Agent, subsequently correcting false detections using the classification Vision Agent. This collaborative interaction effectively combines the MLLM's reasoning capabilities with the object localization strengths of the Vision Agents, resulting in detection outcomes that are more accurate and contextually coherent.

In summary, our contributions are as follows:

\begin{enumerate}
    \item We propose the \textbf{Visual-Linguistic-Agent} (VLA), a novel collaborative framework that utilizes the reasoning and relationship-modeling capabilities of MLLMs to enhance the contextual coherence of object detection.
    \item Within VLA, the MLLM operates as a Linguistic Agent, collaborating with both object detection and classification Vision Agents to filter out false detections by reasoning over spatial relationships and leveraging the classification agent for corrective feedback. This approach maximizes the MLLM's reasoning capabilities while strengthening its localization accuracy, resulting in more contextually consistent detection outcomes.
    \item Extensive experiments on the COCO dataset demonstrate that VLA achieves up to 3\% improvement in AP$_{50:95}$ over state-of-the-art object detection models, underscoring the benefits of integrating linguistic reasoning with visual understanding in object detection and establishing a new benchmark for multimodal capabilities in this field.
\end{enumerate}

% \begin{figure*}[th]
%     \centering
%     \includegraphics[scale=0.47]{figs/paper_figures.pdf}
%     \caption{An example from the proposed \textbf{Visual-Linguistic-Agent (VLA)} paradigm. The Visual Agent (YOLO) detects objects in an image, but mistakenly classifies the moon as an orange. The Linguistic Agent (MLLM) assesses the detection results and, based on its reasoning capabilities, correctly identifies the error. This example highlights the power of integrating visual detection with linguistic reasoning to improve the accuracy and rationality of object detection outcomes..}
%     \label{FIG:att}
% \end{figure*}

\section{Related Work}
\label{sec:formatting}

In this section, we first provide an overview of traditional object detection models and their attempts to model inter-object relationships to reduce false detections. Next, we introduce leading MLLMs, discussing their visual reasoning capabilities. Finally, we compare the strengths and limitations of traditional object detection models and MLLMs with our proposed VLA, highlighting how VLA integrates the strengths of both approaches to achieve more accurate and contextually grounded detection results.

\subsection{Reasoning-Based Modeling in Object Detection}

Reasoning-based modeling in object detection aims to model instance-level contextual relationships among objects to produce more coherent detection results. 

\textit{Implicit modeling} typically uses convolutional neural networks (CNNs) to learn contextual relationships between the visual features of proposed detections, embedding these relationships into proposal features. This approach helps reduce false detections, thus improving the performance of detection models. For instance, Spatial Memory Network (SMN) \cite{chen2017spatial} reassembles object instances into a pseudo "image" to enable object-object context reasoning through a secondary ConvNet. This structure supports parallel processing of image and memory, allowing detected objects to iteratively update memory. Similarly, Relation Networks \cite{hu2018relation} process sets of objects by modeling interactions between their appearance features and geometries, facilitating pairwise relational modeling.

\textit{Explicit modeling} \cite{liu2023cigar, zhao2023rgrn, xu2019spatial} often represents object proposals as nodes, with methods like Graph Convolutional Networks (GCNs) \cite{liu2024rt} or Gated Recurrent Units (GRUs) \cite{chen2017spatial, zulqarnain2024efficient} extracting contextual relationships to improve detection results. Additionally, some methods project visual features of proposals onto pre-defined class text embeddings \cite{zhu2021semantic}. These approaches enhance visual representations by leveraging consistent relational graphs derived from text embeddings, thereby increasing the accuracy and robustness of object detection.

Both implicit and explicit modeling approaches primarily focus on relationships between local objects, often overlooking the broader contextual environment. While these methods embed context within proposal features, they do so in a secondary capacity, as localization and classification tasks remain separate. This separation weakens the role of context, leading to unreasonable detections. In contrast, our proposed VLA leverages MLLMs (acting as a Linguistic Agent) for global visual understanding and natural language reasoning, identifying errors in traditional object detection outputs (produced by the Visual Agent) and utilizing domain-specific Vision Agents to correct them. This approach capitalizes on the MLLM’s reasoning power while enabling the Vision Agents to refine detections, resulting in more accurate and contextually coherent outcomes.

\subsection{AI Agents}

The rapid development of MLLMs \cite{li2024seed, liu2023visual_llava, li2023blip2, touvron2023llama, zhu2023minigpt4, driess2023palme} has led to new research on AI agents. AI agents aim to decompose complex problems into manageable sub-tasks, using Chain-of-Thought (CoT) \cite{wei2022chain} reasoning, where MLLMs serve as the central "brain" that issues commands to specialized models across various domains. For instance, HuggingGPT \cite{shen2024hugginggpt} leverages ChatGPT to coordinate multiple specialized AI models from Hugging Face \cite{jain2022hugging} to produce detailed image captions that describe both visual content and underlying context. ViperGPT \cite{suris2023vipergpt} utilizes code generation models to compose vision-and-language models into structured subroutines, enabling it to generate responses for diverse queries. Compositional Chain-of-Thought (CCoT) \cite{mitra2024compositional} combines large language models with scene graph models to create more accurate scene graphs and enhances the model’s ability to interpret complex visual scenes.

Unlike these AI agents, our VLA integrates the holistic visual understanding and natural language reasoning abilities of MLLMs to identify erroneous detections produced by traditional object detection models. It then employs Classification Vision Agents to correct these errors, resulting in more coherent and contextually rational object detection outcomes. This approach not only makes traditional object detection more robust but also improves the MLLM’s ability to localize objects accurately within images.

\section{Methodology}

The Visual-Linguistic Agent (VLA) is an advanced system designed to enable seamless collaboration between a linguistic agent and specialized, domain-specific visual agents. Section~\ref{sec:statement} examines the limitations of traditional object detection methods, focusing on the causes of unreasonable detections and comparing these with MLLMs in terms of object relationship modeling. Section~\ref{sec:VLA} provides an in-depth introduction to our VLA paradigm, illustrating how diverse visual and natural language agents work together to achieve more contextually coherent object detection.

\subsection{Problem Statement}\label{sec:statement}

\begin{figure}[t]
    \centering
    \includegraphics[width=\linewidth]{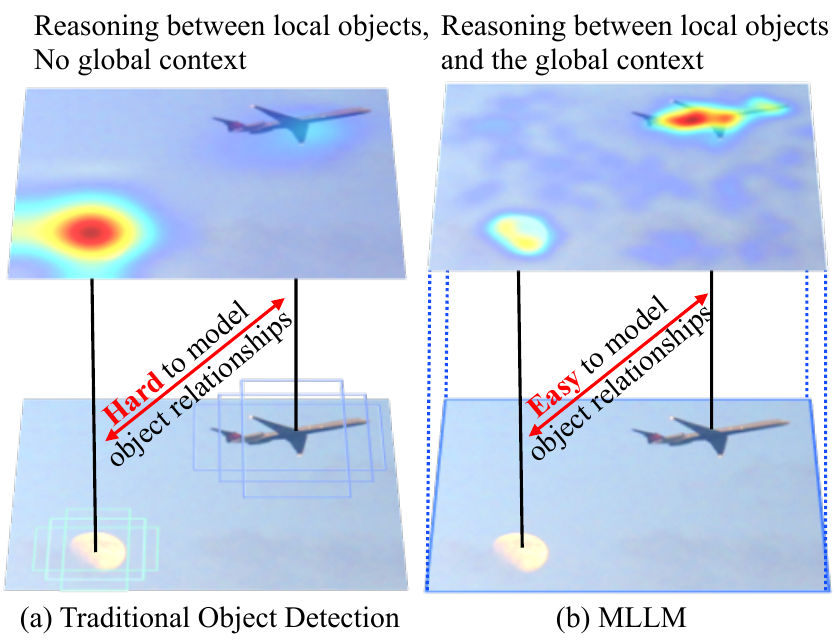}
    \caption{Comparison of traditional object detection (a) and MLLM (b) approaches. Traditional object detection focuses on reasoning among local objects without global context, limiting its ability to model object relationships. In contrast, MLLMs incorporate both local and global context, enabling more comprehensive object relationship modeling.}
    \label{FIG:reason_compare}
\end{figure}

Traditional object detection models generate numerous independent region proposals based on local visual cues. These proposals are then processed by separate branches for position regression and category classification, predicting bounding boxes and categories independently. This approach neglects global contextual relationships, enforcing a direct mapping between local features and category priors without broader contextual coherence. This limitation is particularly evident when the Intersection over Union (IoU) between proposals is low, making it difficult to establish meaningful connections and increasing the likelihood of inconsistent or implausible detections, as shown in Figure \ref{FIG:reason_compare}.

\begin{figure*}[t]
    \centering
    \includegraphics[width=\linewidth]{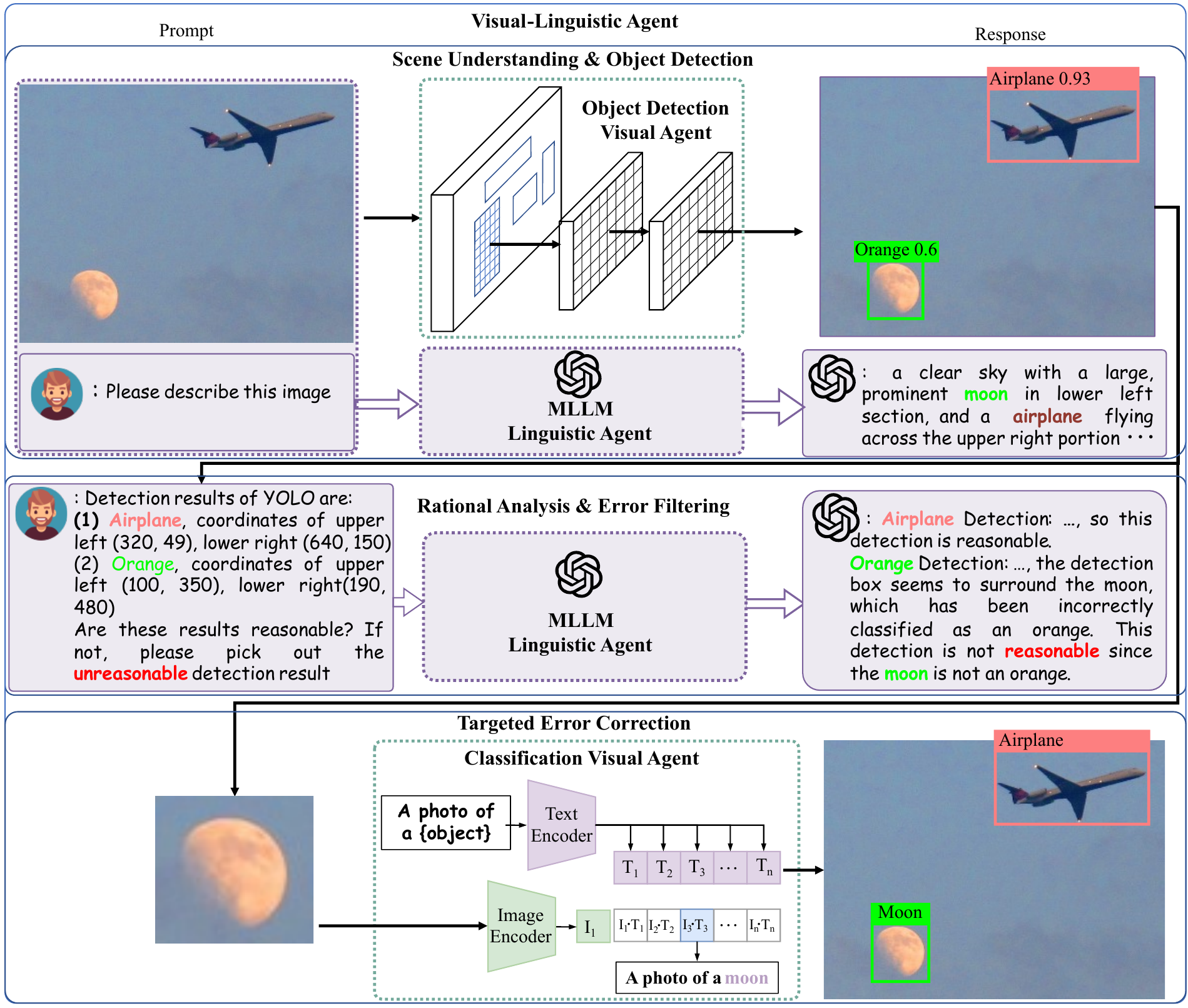}
    \caption{The proposed \textbf{Visual-Linguistic-Agent (VLA)} paradigm. The Visual Agent (e.g., YOLO) detects objects and generates bounding boxes with class labels, which are passed to the Linguistic Agent (MLLM) for reasoning and contextual analysis. Based on the MLLM's assessment, false detections are filtered, and the Classification Visual Agent corrects erroneous detections. This collaboration between agents enhances object detection accuracy and provides more contextually coherent results.}
    \label{FIG:framework}
\end{figure*}

In scenarios where objects are spatially isolated, resulting in low IoU values among objects, the class probability distribution $P(y_i)$ for each object $y_i$ tends to disperse, indicating higher uncertainty. This dispersion can increase false detections, particularly when contextual plausibility is compromised. The model's reliance on local spatial relationships also affects the weighted entropy, $H_{w}(Y)$, defined as:

\begin{equation}
    H_{w}(Y) = - \sum_{i=1}^{n} \left( P(y_i) \cdot w_i \right) \log P(y_i),
\end{equation}

Here, we define $w_i = \frac{1}{1 + \sum_{j \neq i} \text{IoU}(B_i, B_j)}$, where $B_i$ represents the bounding box of the $i$-th object, and $w_i$ approximates 1 for objects with low IoU values, reflecting weak spatial connections and thereby increasing their contribution to $H_w(Y)$. Higher entropy reflects the model’s limited interpretative power, as it cannot establish relationships between objects beyond spatial proximity alone. This limitation often leads to unreasonable detections, where the model lacks the semantic context to discern whether objects logically belong together.

In contrast, MLLMs incorporate a ``global view'', leveraging both visual and linguistic cues to establish coherent semantic relationships based on contextual knowledge. By drawing on large-scale pre-trained knowledge and common-sense reasoning, MLLMs can interpret scene relationships without relying solely on IoU, enabling them to filter out improbable relationships (e.g., “orange in the sky”) in favor of plausible, contextually relevant detections. This capacity for common-sense reasoning results in a more concentrated probability distribution $P(y_i, r_{ij})$, where $r_{ij}$ represents the semantic relationship between objects $y_i$ and $y_j$. The global entropy of the MLLM’s output, $H(Y, R)$, is therefore lower and calculated as:

\begin{equation}
    H(Y, R) = - \sum_{i=1}^{n} \sum_{j \neq i} P(y_i, r_{ij}) \log P(y_i, r_{ij}),
\end{equation}

where $P(y_i, r_{ij})$ represents the MLLM’s refined probability distribution, focused on contextually valid relationships. By concentrating probabilities on plausible combinations and suppressing unlikely predictions, the MLLM achieves a lower entropy $H(Y, R)$, indicating its enhanced capacity for modeling object relationships with reduced uncertainty. This context-aware filtering is particularly valuable in reducing false detections that would otherwise arise in traditional models. 

The difference in entropy between the traditional model’s IoU-based entropy $H_{w}(Y)$ and the MLLM’s global entropy $H(Y, R)$ can be quantified as the information gain (IG):

\begin{equation}
    IG = H_{w}(Y) - H(Y, R).
\end{equation}

A larger $IG$ reflects the extent to which the MLLM reduces uncertainty by leveraging contextual reasoning. This improvement in object relationship modeling, facilitated by the MLLM’s global, context-based perspective, results in more rational and contextually relevant detections. By integrating the MLLM’s semantic reasoning with traditional object detectors, VLA bridges the gap between spatial precision and semantic coherence, enhancing both the accuracy and interpretability of object detection in complex visual scenes. This synergy produces results that more closely align with human intuition and practical scene understanding.

\subsection{Visual-Linguistic Agent}\label{sec:VLA}

As shown in Figure \ref{FIG:framework}, the VLA framework enhances the effectiveness of traditional object detection models by harnessing the natural language reasoning and global scene understanding capabilities of MLLMs. This paradigm operates in three key stages: 1) global scene understanding and object detection, 2) rational analysis and error filtering using MLLMs, and 3) targeted error correction. The following sections detail the design of each phase.

\subsubsection{Scene Understanding and Object Detection}

In the first stage, \textbf{Scene Understanding and Object Detection}, VLA combines \textit{scene comprehension} with \textit{object detection} through the collaboration between the \textbf{Visual Agent} and the MLLM, acting as the \textbf{Linguistic Agent}.

The \textbf{Visual Agent} (e.g., YOLO) processes an input image to generate bounding boxes and class labels for detected objects. The output from the Visual Agent includes:
\begin{itemize}
    \item \textbf{Input}: Image
    \item \textbf{Output}: Object detection results with bounding box coordinates and class labels, such as:
    \begin{itemize}
        \item Airplane, coordinates: upper-left (320, 49), lower-right (640, 150)
        \item Orange, coordinates: upper-left (100, 350), lower-right (190, 480)
    \end{itemize}
\end{itemize}

Simultaneously, the \textbf{Linguistic Agent} (MLLM) generates a contextual \textbf{image caption} based on visual features to summarize the scene. For example, if the image shows an airplane flying in the sky with the moon visible, as illustrated in Figure \ref{FIG:framework}, the caption might be:
\begin{quote}
    ``The image shows an airplane flying across the sky with the moon in the background''.
\end{quote}

The detection results from the \textbf{Visual Agent} are then converted into a text prompt and passed to the \textbf{Linguistic Agent} for further analysis. An example prompt might be:
\begin{quote}
    ``The object detector identified the following objects: 
    \begin{itemize}
        \item Airplane, coordinates: (320, 49), (640, 150)
        \item Orange, coordinates: (100, 350), (190, 480)
    \end{itemize}
    Are these results reasonable based on the scene context?''
\end{quote}

This prompt serves as the input for the \textbf{Linguistic Agent} in the next stage, establishing a seamless integration between the initial detection results and the rational analysis that follows.

\subsubsection{Rational Analysis and Error Filtering}

In this stage, the \textbf{Linguistic Agent} processes the prompt generated in the previous step, evaluating the reasonableness of each detection based on its understanding of the scene and common-sense knowledge. The MLLM compares the detected objects against the contextual information derived from the image caption and spatial relationships among objects.

For instance, if the detection results include an object classified as an ``orange'' with coordinates that match the position of the moon, the \textbf{Linguistic Agent} can infer that this classification is likely incorrect based on the scene context (as the moon would not logically be identified as an orange). The output of this stage is a refined set of detection results, where false detections are flagged for correction.

\begin{itemize}
    \item \textbf{Input}: Text prompt with detection results from the Visual Agent.
    \item \textbf{Output}: Flagged results identifying false detections. For example:
    \begin{quote}
        ``Airplane detection is reasonable. 
        Orange detection is unreasonable, as the object is likely the moon''.
    \end{quote}
\end{itemize}

These flagged detection results are then passed as input to the next stage, where targeted error correction is performed.

\subsubsection{Targeted Error Correction}

In the final stage, \textbf{Targeted Error Correction}, the system addresses flagged detections identified by the \textbf{Linguistic Agent}. False detections are corrected by engaging a \textbf{Classification Visual Agent}, which performs fine-grained classification on the specified regions. As shown in Figure \ref{FIG:framework}, the region where the ``orange'' was misclassified is sent to the classification model for correction, resulting in an updated label:
\begin{quote}
    Region (100, 350), (190, 480): reclassified as ``moon''.
\end{quote}

This final step integrates corrections from the classification model with the capabilities of the detection model, ensuring the detection results are both accurate and contextually relevant. The collaboration between the \textbf{Linguistic Agent} and the \textbf{Visual Agents} optimizes the MLLM’s reasoning power while enhancing object detection accuracy, resulting in more coherent and rational detection outcomes.

\begin{figure*}
    \centering
    \includegraphics[width=\linewidth]{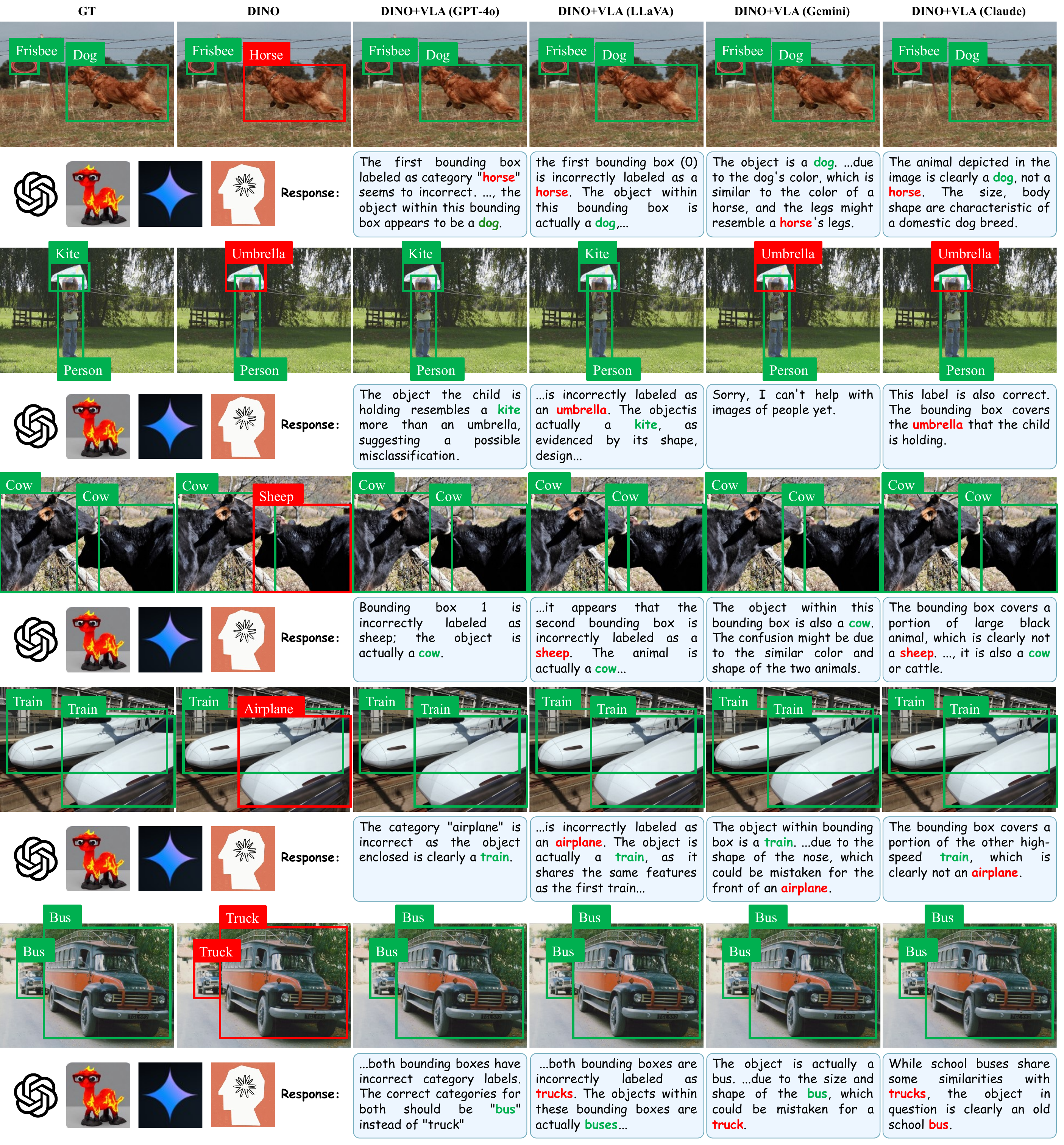}
    \caption{Examples of error correction by VLA using DINO with MLLMs as the Linguistic Agent. The figure highlights common misclassifications corrected by VLA, such as identifying a dog instead of a horse, recognizing a kite instead of an umbrella, and correctly labeling a cow instead of a sheep. These corrections demonstrate VLA's ability to refine object detection outputs by leveraging linguistic reasoning to resolve ambiguous or erroneous labels.}
    \label{FIG:results}
\end{figure*}

\section{Experiments}

\subsection{Experimental Settings}

The Visual-Linguistic-Agent framework is evaluated using a range of MLLMs as the Linguistic Agent, including GPT-4o \cite{kevian2024capabilities}, Claude 3.5 Sonnet \cite{kevian2024capabilities}, LLaVA \cite{kevian2024capabilities}, and Gemini 1.5 \cite{kevian2024capabilities}. Each model offers distinct strengths in scene understanding and contextual reasoning. For object detection, we employ Faster R-CNN \cite{ren2016faster}, YOLOX \cite{tsai2024sw}, YOLOv11 \cite{khanam2024yolov11}, DETR \cite{carion2020end}, and DINO \cite{zhangdino} as the Visual Agents to generate initial bounding boxes and class labels. Additionally, CLIP is used as a Classification Visual Agent to refine false detections and verify object classifications, leveraging its strong text-image alignment capabilities. The Linguistic and Visual Agents exchange information using JSON format for structured data transfer, with final results saved in COCO-style JSON files for performance evaluation.

Experiments are conducted on the COCO dataset, a comprehensive benchmark containing diverse object categories, sizes, and backgrounds, making it ideal for assessing detection and correction performance. Evaluation metrics include Detection Metrics such as mean Average Precision (mAP) \cite{vo2024robust}, with AP$_{50:95}$, AP$_{50}$, AP$_{75}$, AP$_\text{s}$ (small objects), AP$_\text{m}$ (medium objects), and AP$_\text{l}$ (large objects) as key indicators of detection performance across categories. Additionally, Correction Rate is introduced to assess the VLA framework’s accuracy in correcting unreasonable labels, focusing on false detections and classification accuracy. This setup provides a thorough analysis of each model configuration within the VLA paradigm, revealing insights into the strengths of different MLLM and object detection model combinations.

\subsection{Quantitative Results}

We evaluate the performance improvements achieved by integrating the proposed VLA framework with various object detection models on the COCO dataset. Table \ref{tab:vla_performance} presents the results across metrics. The integration of VLA consistently enhances performance across all models. For instance, Faster R-CNN's AP$_{50:95}$ improved from 37.4 to 40.1 with VLA, while YOLOX’s AP$_{50:95}$ increased from 40.3 to 42.7. DINO, the highest-performing model, achieved an AP$_{50:95}$ of 49.5, which further rose to 52.1 when integrated with VLA. These improvements highlight VLA's ability to enhance detection accuracy by leveraging contextual reasoning, particularly in complex scenes and for small objects where object relationships are critical.

\begin{table}
    \centering
    \caption{Performance (\%) of different object detection models on the COCO dataset with and without VLA integration. }
    \label{tab:vla_performance}
    \resizebox{0.49\textwidth}{!}{
    \begin{tabular}{lcccccc}
        \toprule
        \textbf{Model} & \textbf{AP$_{50:95}$} & \textbf{AP$_{50}$} & \textbf{AP$_{75}$} & \textbf{AP$_\text{s}$} & \textbf{AP$_\text{m}$} & \textbf{AP$_\text{l}$} \\
        \midrule
        Faster R-CNN & 37.4 & 58.1 & 40.4 & 21.2 & 41.0 & 48.1 \\
        Faster R-CNN+VLA & 40.1 & 63.2 & 43.1 & 24.8 & 43.7 & 51.3 \\
        YOLOX & 40.3 & 59.1 & 43.4 & 23.5 & 44.5 & 53.1 \\
        YOLOX+VLA & 42.7 & 63.4 & 45.8 & 25.2 & 46.7 & 56.7 \\
        YOLO11 & 48.5 & 62.3 & 52.8 & 28.6 & 53.9 & 66.6 \\
        YOLO11+VLA & 49.8 & 64.1 & 54.2 & 29.1 & 54.9 & 68.3 \\
        DETR & 39.7 & 60.3 & 41.4 & 17.2 & 43.2 & 59.2 \\
        DETR+VLA & 41.4 & 63.3 & 43.2 & 18.8 & 45.1 & 61.1 \\
        DINO & 49.5 & 67.3 & 53.9 & 32.2 & 52.7 & 64.2 \\
        DINO+VLA & 52.1 & 71.2 & 56.9 & 36.1 & 55.2 & 66.9 \\
        \bottomrule
    \end{tabular}}
\end{table}

The results in Table \ref{tab:vla_performance} further demonstrate that VLA integration improves the detection of medium and large objects across models. By refining object detection outputs through context-aware analysis, VLA provides a notable increase in precision across a range of challenging scenarios. Moreover, this paradigm effectively maximizes the MLLM's reasoning capabilities, while domain-specific Vision Agents carry out tasks delegated by the Linguistic Agent, resulting in more accurate and contextually coherent detections.

\subsection{Visual Reasoning Capabilities of MLLMs}

Table \ref{tab:dino_vla_performance} and Figure \ref{FIG:results} present the performance of DINO with and without VLA integration, using various MLLMs as the Linguistic Agent. The results indicate that VLA integration consistently enhances DINO's detection metrics, with significant improvements in AP$_{50:95}$, AP$_{50}$, and AP$_{75}$. Notably, DINO+VLA (GPT-4o) achieves the highest AP$_{50:95}$ at 52.1, an improvement of 2.6 points over the baseline DINO model's AP$_{50:95}$ of 49.5. Similarly, DINO+VLA (LLaVA) increases AP$_{50:95}$ by 2.4 points, reaching 51.9. GPT-4o also yields substantial gains in AP$_{50}$ (+3.9) and AP$_{75}$ (+3.0), showcasing its strong contextual reasoning capabilities that enhance detection accuracy, particularly for small and medium objects.

 \begin{table}
    \centering
    \caption{Detection Performance (\%) of DINO with and without VLA integration using different MLLMs as the Linguistic Agent.}
    \label{tab:dino_vla_performance}
    \resizebox{0.49\textwidth}{!}{
    \begin{tabular}{lcccccc}
        \toprule
        \textbf{Model} & \textbf{AP$_{50:95}$} & \textbf{AP$_{50}$} & \textbf{AP$_{75}$} & \textbf{AP$_\text{s}$} & \textbf{AP$_\text{m}$} & \textbf{AP$_\text{l}$} \\
        \midrule
        DINO & 49.5 & 67.3 & 53.9 & 32.2 & 52.7 & 64.2 \\
        DINO+VLA (Claude) & 51.7 & 70.7 & 55.9 & 36.8 & 54.7 & 65.6 \\
        DINO+VLA (Gemini) & 51.6 & 70.5 & 55.5 & 35.3 & 54.4 & 65.3 \\
        DINO+VLA (LLaVA) & 51.9 & 71.1 & 56.7 & 36.2 & 55.0 & 67.1 \\
        DINO+VLA (GPT-4o) & 52.1 & 71.2 & 56.9 & 36.1 & 55.2 & 66.9 \\
        \bottomrule
    \end{tabular}}
\end{table}

VLA integration also provides notable improvements in detecting smaller objects (AP$_\text{s}$), with all MLLM configurations showing gains. For example, DINO's AP$_\text{s}$ improves by 4.6 points with Claude, while LLaVA and GPT-4o also demonstrate significant boosts, underscoring VLA's effectiveness in refining object detection for smaller, challenging objects. Overall, as shown in Figure \ref{FIG:results}, the choice of MLLM impacts VLA's effectiveness. While all configurations yield performance enhancements, GPT-4o and LLaVA stand out, demonstrating superior contextual reasoning and refinement capabilities within the VLA framework.

\subsection{Ablation Study}

\textbf{Effectiveness of VLA Components:} Table \ref{tab:error_correction} evaluates the error correction capabilities of the DINO model with and without the integration of the Linguistic Agent (LA) and VLA. Metrics include Error Detections (ED, total erroneous detections identified), Corrected Detections (CD, number of errors corrected by the framework), and Corrected Rate (CR, percentage of corrected errors). The baseline DINO model shows a Corrected Rate of 0\%, indicating its inability to self-correct erroneous detections. With the integration of LA, the CR increases to 44.9\%, highlighting that LA’s reasoning capabilities can partially assist in identifying and correcting errors. However, due to limited local context, LA alone is less effective in complex scenarios.

The DINO+VLA configuration achieves the highest improvement, with a Corrected Rate of up to 75.0\%. This result emphasizes the advantage of combining the MLLM's global reasoning with the Classification Visual Agent’s local object understanding. By integrating VLA, the framework not only filters out unreasonable detections but also enhances the MLLM's ability to process local visual information, enabling more comprehensive error correction. Table \ref{tab:ap_performance} further illustrates detection performance, with DINO+VLA achieving the best results across all AP metrics, particularly AP$_{50:95}$ and AP$_{75}$, demonstrating how VLA enhances detection accuracy by combining global context reasoning with precise local object recognition.

In conclusion, while LA integration improves the DINO model’s error correction capability through global contextual reasoning, it lacks the precision required for local object detection. The VLA framework effectively merges the MLLM’s global reasoning capabilities with enhanced local visual understanding, achieving the highest correction rates and detection performance. This underscores the importance of combining linguistic and visual agents to achieve reliable and contextually coherent object detection.

\begin{table}
    \centering
    \caption{Error correction capabilities of DINO with and without the integration of LA and VLA. ED represents the number of label prediction errors in bounding boxes generated by DINO.}
    \label{tab:error_correction}
    
    \begin{tabular}{lccc}
        \toprule
        Model & ED & CD & CR \\
        \midrule
        DINO & 1327 & 0 & 0\% \\
        DINO+LA (GPT-4o) & 1327 & 597 & 44.9\% \\
        DINO+VLA (Claude) & 1327 & 982 & 74.0\% \\
        DINO+VLA (Gemini) & 1327 & 979 & 73.7\% \\
        DINO+VLA (LLaVA) & 1327 & 990 & 74.6\% \\
        DINO+VLA (GPT-4o) & 1327 & 996 & 75.0\% \\
        \bottomrule
    \end{tabular}
\end{table}

\begin{table}
    \centering
    \caption{AP performance (\%) of the DINO model with and without integration of LA and VLA.}
    \label{tab:ap_performance}
    \resizebox{0.49\textwidth}{!}{
    \begin{tabular}{lcccccc}
        \toprule
        Model & AP$_{50:95}$ & AP$_{50}$ & AP$_{75}$ & AP$_\text{s}$ & AP$_\text{m}$ & AP$_\text{l}$ \\
        \midrule
        DINO & 49.5 & 67.3 & 53.9 & 32.2 & 52.7 & 64.2 \\
        DINO+LA & 51.6 & 69.8 & 52.3 & 35.2 & 54.2 & 65.0 \\
        DINO+VLA & 52.1 & 71.2 & 56.9 & 36.1 & 55.2 & 66.9 \\
        \bottomrule
    \end{tabular}}
\end{table}

\section{Conclusion}

We presented the \textbf{Visual-Linguistic-Agent} (VLA), a novel paradigm designed to enhance traditional object detection models by leveraging the reasoning and relational capabilities of multimodal large language models (MLLMs). Within the VLA framework, the MLLM serves as a Linguistic Agent that collaborates with both an object detection Vision Agent and a classification Vision Agent. This setup enables the MLLM to generate contextual captions, assess detection outputs, and make informed corrections, integrating its global reasoning capabilities with the precise localization strengths of the Vision Agents. Consequently, VLA achieves detection outcomes that are not only more accurate but also more contextually coherent. Experimental results on the COCO dataset demonstrate that VLA effectively reduces unreasonable detections by incorporating contextual understanding.

\textbf{Limitations:} While VLA is effective in addressing clear labeling errors and enhancing contextual coherence, certain complex detection issues remain challenging. For instance, handling redundant bounding boxes on a single object, which may occasionally bypass non-maximum suppression, requires further exploration. Addressing these nuanced cases could provide additional refinements to VLA’s performance in future work.

{
    \small
    \bibliographystyle{ieeenat_fullname}
    \bibliography{main}
}


\end{document}